%% file: main.tex
\documentclass[runningheads]{llncs}
\usepackage{graphicx}
\usepackage{multirow}
\usepackage{amssymb}
\usepackage{wrapfig}
\usepackage{floatflt}
\usepackage{amsmath}
\usepackage{xspace}

\newcommand{\name}[0]{MASSM\xspace}
\begin{document}
\title{\name: An End-to-End Deep Learning Framework for Multi-Anatomy Statistical Shape Modeling Directly From Images}

\author{Janmesh Ukey \and Tushar Kataria \and Shireen Y. Elhabian}
%
%
\authorrunning{
J Ukey et al.}

\titlerunning{MASSM: Multi-Anatomy Statistical Shape Modeling}
%
\institute{Kahlert School of Computing, University of Utah \and
Scientific Computing and Imaging Institute, University of Utah \\ 
\{janmesh,tushar.kataria,shireen\}@sci.utah.edu }

\maketitle              
\input{abstract}
\input{introduction}
\input{method}
\input{results}
\input{conclusion}


\bibliographystyle{splncs04.bst}
\bibliography{ref.bib}
\newpage

\end{document}

%% file: abstract.tex
\begin{abstract}
Statistical Shape Modeling (SSM) effectively analyzes anatomical variations within populations but is limited by the need for manual localization and segmentation, which relies on scarce medical expertise.
%
Recent advances in deep learning have provided a promising approach that automatically generates statistical representations (as point distribution models or PDMs) from unsegmented images. Once trained, these deep learning-based models eliminate the need for manual segmentation for new subjects.
%
Most deep learning methods still require manual pre-alignment of image volumes and bounding box specification around the target anatomy, leading to a partially manual inference process.
Recent approaches facilitate anatomy localization but only estimate population-level statistical representations and cannot directly delineate anatomy in images. Additionally, they are limited to modeling a single anatomy.
%
%
%
We introduce \name, a novel end-to-end deep learning framework that simultaneously localizes multiple anatomies, estimates population-level statistical representations, and delineates shape representations directly in image space.
%
Our results show that \name, which delineates anatomy in image space and handles multiple anatomies through a multitask network, provides superior shape information compared to segmentation networks for medical imaging tasks. Estimating Statistical Shape Models (SSM) is a stronger task than segmentation, as it encodes a more robust statistical prior for the objects to be detected and delineated. \name allows for more accurate and comprehensive shape representations, surpassing the capabilities of traditional pixel-wise segmentation.

\keywords{Multi-Anatomy Networks \and Deep Learning \and Statistical Shape Modeling \and Anatomy Detection \and Localization}
\end{abstract}

%% file: introduction.tex
\section{Introduction}
Statistical Shape Modeling (SSM) is a powerful technique for quantifying and studying variations in anatomical forms. 
SSM has demonstrated its invaluable utility across a spectrum of biomedical and clinical applications, exemplified by studies and applications such as morphometrics \cite{doi:10.1177/10711007241231981}, orthopedics \cite{atkins2019two,almalki2023statistical,atkins2023correspondence,roda2023above}, growth modeling \cite{jeffery2023postnatal,atkins2017quantitative} and surgical planning \cite{nguyen2023feasibility}. SSM involves deriving parameters that can represent anatomy in the context of a population. 
A key technique entails using \textit{landmarks} or \textit{correspondence points} defined by their consistent anatomical correspondences across different subjects in a population. 
%
%
Traditionally, generating such correspondences/landmarks has been a manual and time-consuming process, heavily reliant on radiologist/medical expertise as it requires precise delineations and annotation of distinct features on the anatomy of interest. 

Existing SSM methods \cite{davies2002minimum,sarkalkan2014statistical,cates2017shapeworks,durrleman2014morphometry,vicory2018slicersalt} predominantly employ the automated placement of dense correspondences, known as point distribution models (PDMs) on anatomical structures. These correspondences placements is achieved through different optimization-based frameworks such as entropy minization for particles placed on segmented images obtained manually. 
%
%
Even though such advancements reduce the need for expert manual identification of unique shape features, they still entail expert-guided workflows for manual anatomy segmentation by medical professionals, shape registration, optimization of population-level shape representations, and extensive hyper parameter tuning. Despite technological progress, these steps remain time-intensive and costly, requiring considerable expert involvement.

Deep learning methods have significantly improved SSM workflows, providing an automated approach to deriving statistical representations of anatomies directly from unsegmented images \cite{bhalodia2018deep,bhalodia2018deepssm,adams2020uncertain,adams2022images,tao2022learning,huang2017temporal,milletari2017integrating,xie2016deepshape,zheng20153d,raju2022deep,karanam2023adassm,bhalodia2024deepssm,adams2024weakly,iyer2024scorp}. Once properly trained, these deep learning models eliminate the need for manual segmentation for new subjects, thereby streamlining the estimation of PDMs on new subjects during deployment. 
%
However, these approaches still necessitate image pre-processing steps, such as anatomy localization and rigid alignment. This involves cropping the images around the area of interest and aligning them with a reference shape during both the training and inference phases. 
%
This prerequisite hinders the deployment of these methods as a fully automated alternative to traditional, optimization-based SSM techniques.

Ukey et al. \cite{ukey2023localization} has addressed the challenges of automated anatomy localization, rigid pose alignment, and SSM prediction, proposing an end-to-end model. However, they primarily estimate population-level statistical representations of anatomies. While the integration of localization and alignment helps in regularizing the SSM learning process, it also imposes limitations on the method’s versatility in delineating anatomy within the image space and confines the analysis to a single anatomy.
To the best of our knowledge, no existing method simultaneously provides population-level shape statistics and image-level anatomy delineation. Moreover, existing approaches are designed to operate exclusively for a single anatomy, necessitating the training of a separate model for each specific anatomical structure. This requirement limits scalability, particularly for on-demand image-based diagnostics. 

In this paper, we introduce \name, a multi-anatomy deep learning framework that simultaneously detects multiple anatomies within an 3D-image, accurately delineates each anatomy within the image to extract the corresponding shape representation, and estimates the corresponding population-level statistical shape representation of each detected anatomy. 
\name demonstrates better results than the models trained on single anatomy individually. Additionally \name is scalable to accommodate any number of anatomies, constrained only by available computational resources. Furthermore, our image-level shape representation is able to provide a better surface-to-surface estimate than the segmentation methods. Unlike segmentation tasks that involve pixel-wise classification, Statistical Shape Modeling (SSM) encodes a stronger statistical prior for the objects being detected and delineated, resulting in more accurate and robust shape representations.
%
 The main contributions of this paper are:
 \begin{itemize}
     \item We propose a novel end-to-end deep learning framework for simultaneous multi-anatomy detection and statistical shape modeling directly from unsegmented images.
     \item The proposed method is a unified framework that offers both image-level and population-level shape representation in form of Point Distribution Models (PDMs).
 \end{itemize}

%% file: method.tex
\section{Methods}
Given a dataset of $N$ samples, we denote the 3D images as $\{\mathbf{I}_n\in \mathbb{R}^{H \times W \times D} \}_{n=1}^N$. MASSM generates both population-level and image-level shape representations in the form of Point Distribution Model (PDM) for the maximum of $K$ anatomies present in the images $\mathbf{I}_n$.


For shape analysis, we characterize shape as the residual information after eliminating global alignment differences across samples within a given cohort. The PDM representing population-level statistical shape information after removing global alignment, is termed \textit{world correspondences}, and denoted by $\mathbf{W}_n^{k}$. $\mathbf{W}_n^{k}={w_1, ... w_M}$, is ordered set of 3d points, where $w_m \in R^3$, represents the correspondence points, where $k$ represents the $k$-th anatomy in the image. The process of eliminating global alignment differences involves aligning all shapes to a template shape. We select the medoid shape (represented by $\overline{\mathbf{W}}^{k}$) as the template; the mediod for each anatomy type $k$ is selected by first computing the average segmentation of the training samples and then selecting the shape nearest to it from training set. 

The PDM capturing shape information in image space, i.e., before eliminating global alignment is referred to as \textit{local correspondences}, denoted by $\mathbf{L}_n^{k}$, is ordered set of 3d points in image space, $\mathbf{L}_n^{k} = {l_1, ... l_M}$ where $l_m \in R^3$. $\overline{\mathbf{L}}^{k}$ represents the local correspondences for the mediod shape for an anatomy $k$.
\subsection{MASSM}
\begin{figure}[!htb]
\vspace{-1em}
  \centering
  \includegraphics[width=0.8\linewidth]{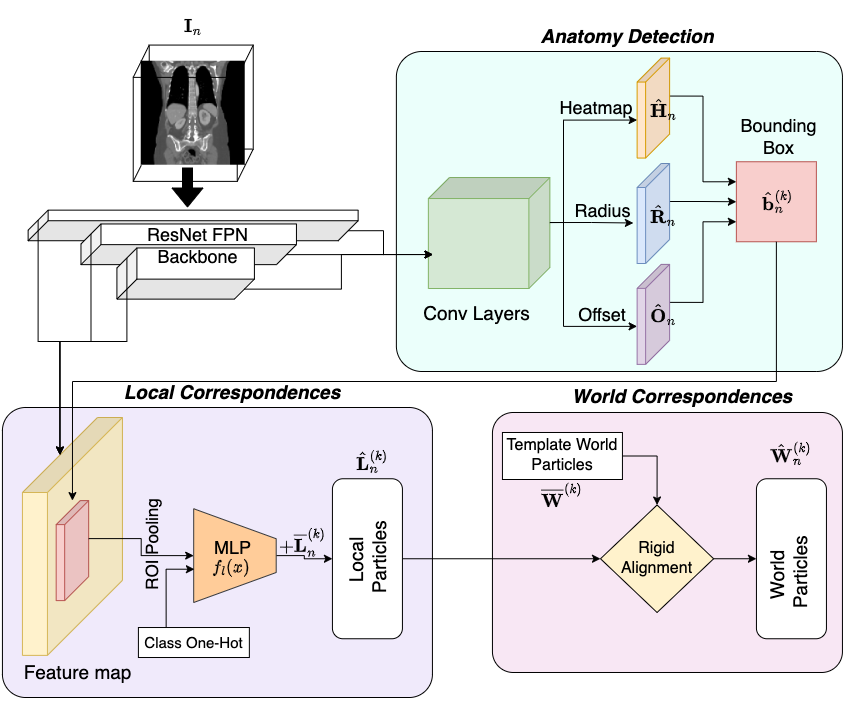}
  \caption{\textbf{Multi-Anatomy Statistical Shape Model (\name).} Block diagram of the proposed end-to-end method to obtained statistical shape representation of multiple anatomies simultaneously. The proposed model has three networks; (a) \textit{Anatomy Detection}, which extracts different anatomies of interest; (b) \textit{Local Correspondences}, which predicts local particle correspondences and; (c) \textit{World Correspondences}, which predicts world correspondences.}
  \label{arch}
  \vspace{-2em}
\end{figure}

MASSM generates 3D local and world correspondences for different anatomies present in $\mathbf{I}_n$. 
The proposed architecture has three sub-networks: (1) Anatomy  Detection Block, (2) 3D local correspondences predictor, and (3) 3D world correspondences predictor. The block diagram of the full framework is shown in Figure \ref{arch}. 
\subsection{Anatomy Detection Block}

The Anatomy Detection block follows CenterNet \cite{zhou2019objects}  architecture with a 3D-ResNet-101 FPN backbone. Let $(c_x^{n,k}, c_y^{n,k}, c_z^{n,k)}$ be the ground truth center of anatomy of class $k \in (1,2...K)$ in image $\mathbf{I}_n$. Let  $(r_x^{n,k}, r_y^{n,k}, r_z^{n,k})$ be the radius from center that forms the bounding box over anatomy $k$. 

The detection model utilizes a fusion of multi-resolution features to produce three distinct maps at strided output dimensions with a stride of $R$. The stride $R$ is a hyperparameter that balances model efficiency and detection accuracy. Typically, a slightly lower resolution heatmap suffices for accurately predicting the centers and dimensions of objects while maintaining model efficiency. The CenterNet model predicts a center heatmap map $\hat{\mathbf{H}}_n \in \mathbb{R}^{K \times \frac{H}{R} \times \frac{W}{R} \times \frac{D}{R}}$, a radius map $\hat{\mathbf{R}}_n \in \mathbb{R}^{3 \times \frac{H}{R} \times \frac{W}{R} \times \frac{D}{R}}$ and an offset map $\hat{\mathbf{O}}_n \in \mathbb{R}^{3 \times \frac{H}{R} \times \frac{W}{R} \times \frac{D}{R}}$ (for our experiments, we set $R=4$ based on cross-validation results).

The center heat map predicts the center of the anatomy. However these are predicted on strided output dimensions, so these coordinates do not correspond to real centers in the image. The offset map provides the offset from the center in the heatmap to the real center. We use the predicted heatmap center $(h_x^{n,k}, h_y^{n,k}, h_z^{n,k})$ and predicted offset $(o_x^{n,k}, o_y^{n,k}, o_z^{n,k})$ to extract the true center of the anatomy using the equations below:
\begin{align}
    \widehat{c_x}^{n,k} &= h_x^{n,k} \times R + o_x^{n,k}\\
    \widehat{c_y}^{n,k} &= h_y^{n,k} \times R + o_y^{n,k}\\
    \widehat{c_z}^{n,k} &= h_z^{n,k} \times R + o_z^{n,k}
\end{align}
\noindent
We also use the above predicted heatmap centers $(h_x^{n,k}, h_y^{n,k}, h_z^{n,k})$ to extract the corresponding bounding box radii $(\widehat{r_x}^{n,k}, \widehat{r_y}^{n,k}, \widehat{r_z}^{n,k})$ from the radius map, where each channel of the radius map gives the radius or the distance from center to the end of bounding box.
The anatomy detection block is trained using focal loss for heatmap ($\mathcal{L}_h$) and the offset map ($\mathcal{L}_o$), and masked MSE loss for the radius map ($\mathcal{L}_r$).
\begin{align}
    \mathcal{L}_h =& \cfrac{1}{N} \sum 
    \begin{cases}
    (1-\hat{\mathbf{H}}_n)^\alpha\log{(\hat{\mathbf{H}}_n)}, & \text{if } \mathbf{H}_n = 0 \\
    (1-\mathbf{H}_n)^\beta(\hat{\mathbf{H}}_n)^\alpha\log{(1-\hat{\mathbf{H}}_n)}, & \text{otherwise}
    \end{cases}\\
    \mathcal{L}_r =& \cfrac{1}{N} \sum \lVert \hat{\mathbf{R}}_n - \mathbf{R}_n\rVert^2\\
    \mathcal{L}_o =& \cfrac{1}{N} \sum \cfrac{\lVert\mathbf{O}_n - \hat{\mathbf{O}}_n \rVert^2} {1+\exp(a\cdot(c-\lVert \mathbf{O}_n - \hat{\mathbf{O}}_n \rVert))}
\end{align}
\noindent here, $(\alpha, \beta)$ and $(a, c)$ are hyperparameters of $\mathcal{L}_h$ and, $\mathcal{L}_o$ respectively. For all our experiments we set $\alpha=3$, $\beta=4$, $a=10$ and $c=0.2$, which were determined using cross-validation.

\subsection{Local Correspondences}
Local correspondences provide the shape representation of delineation of the anatomy in image space and hence can be further used for supervising other processes like segmentation or registration.

For $k^{th}$ anatomy in the full 3D image $\mathbf{I}_n$, let $(\widehat{c_x}^{n,k}, \widehat{c_y}^{n,k}, \widehat{c_z}^{n,k})$ and $(\widehat{r_x}^{n,k},$ $ \widehat{r_y}^{n,k}, \widehat{r_z}^{n,k})$ be the predicted center and radius of the bounding box $\mathbf{b}_{n,k}$, respectively. For each bounding box detection, the corresponding multi-resolution ROI features are extracted from the FPN backbone. ROI pooling then fuses these features into a single 1D vector. A one-hot vector corresponding to class $k$ ($k^{th}$ position in vector of size $K$ is 1 and all others are 0) is appended to create a final feature vector for an anatomy of interest. This class-appended feature vector is given as input to an MLP layer ($f_l(x)$) to predict the displacement for each local particle. 

The center of local correspondences of the template shape $\overline{\mathbf{L}}^{k}$ is first aligned to the predicted center $(\widehat{c_x}^{n,k}, \widehat{c_y}^{n,k}, \widehat{c_z}^{n,k})$, resulting in $\overline{\mathbf{L}}_n^{k}$. Subsequently, our network $f_l(x)$ predicts the displacement of each correspondence point from $\overline{\mathbf{L}}_n^{k}$ to the actual local correspondences. The predicted local correspondences $\hat{\mathbf{L}}_n^{k}$ is be defined by, 
\begin{align}
    \hat{\mathbf{L}}_n^{k} = \overline{\mathbf{L}}_n^{k} + d \tanh{f_l(x)}
\end{align}
where, $d=2 \cdot\max(\widehat{r_x}^{n,k}, \widehat{r_y}^{n,k}, \widehat{r_z}^{n,k})$, gives the maximum dimension of the bounding box. This component of the framework optimizes a Focal Loss ($\mathcal{L}_l$) between the estimated and the ground truth local particles. Here, $a, c$ are the hyper parameters of $\mathcal{L}_l$, which are set to $a=10$ and $c=0.2$ for all experiments and were found using cross-validation.
\begin{align}
    \mathcal{L}_l =& \cfrac{1}{N} \sum_n \sum_k \cfrac{\lVert\widehat{\mathbf{L}^{k}}_n - \mathbf{L^{k}}_n\rVert^2}{1+\exp(a\cdot(c-\lVert \widehat{\mathbf{L^k}}_n - \mathbf{L^k}_n \rVert))}
\end{align}

\subsection{World Correspondences}
For each anatomy detection, we employ a rigid registration to align corresponding predicted local particles $\hat{\mathbf{L}}_n^{(k)}$ to the template world correspondences for an anatomy index by $k$ $(\overline{\mathbf{W}}^{(k)})$. 

We used Procrustes alignment, which involves centering both $\hat{\mathbf{L}}_n^{(k)}$ $(\overline{\mathbf{W}}^{(k)})$ around their respective centroids, achieved by computing and subtracting the mean of each set. Then computing the covariance matrix of these centered points and utilizing Singular Value Decomposition (SVD) to derive the matrices $U$, $S$, and $V^T$, which facilitate the computation of the optimal rotation matrix($\mathcal{R}$). Subsequently, the optimal translation vector($\mathbf{T}$) is computed to align the centroid of the rotated $\hat{\mathbf{L}}_n^{k}$ with that of $(\overline{\mathbf{W}}^{k})$. Applying the computed rotation matrix and translation vector to $\hat{\mathbf{L}}_n^{k}$ results in the predicted world correspondences $\hat{\mathbf{W}}_n^{k}$.

\begin{align}
    \hat{\mathbf{W}}_n^{(k)} = \mathcal{T} \cdot \hat{\mathbf{L}}_n^{k}
\end{align}
where $\mathcal{T}=(\mathcal{R},\mathbf{T})$ is transformation matrix found using procrustes alignment algorithm. As Eq 9 is differentiable, we calculate the Focal Loss ($\mathcal{L}_w$) between the estimated and ground-truth world particles. 
\begin{align}
    \mathcal{L}_w =& \cfrac{1}{N} \sum_n \sum_k \cfrac{\lVert\widehat{\mathbf{W}^{k}}_n - \mathbf{W^{k}}_n\rVert^2}{1+\exp(a\cdot(c-\lVert \widehat{\mathbf{W^k}}_n - \mathbf{W^k}_n \rVert))}
\end{align}
where $a, c$ are the hyperparameters of $\mathcal{L}_w$, which are set to $a=10$ and $c=0.2$ for all experiments and were found using cross-validation..

\subsection{End-to-End MASSM}
Anatomy detection, local correspondences and world correspondences networks are connected as shown in Figure \ref{arch}. The final loss used to train end-to-end MASSM is given by 
\begin{align}
    \mathcal{L} = \lambda_h \mathcal{L}_h + \lambda_r \mathcal{L}_r + \lambda_o \mathcal{L}_o + \lambda_l \mathcal{L}_l + \lambda_w \mathcal{L}_w
\end{align}

\noindent
where  $\lambda_h,  \lambda_r,  \lambda_o,  \lambda_l,  \lambda_w$ are hyperparameters.

%% file: results.tex
\section{Results and Discussion} \label{results}
\textbf{Training Details.}
The model is trained in a phase-wise manner. The anatomy detection block is trained first with the initial hyperparameter values in equation (1) set to $\lambda_h=1$, $\lambda_r=0.01$ and $\lambda_o=1$. After 20 epochs, $\lambda_l$ is slowly increased with step size of $0.2$ until it reaches the max value of $2$ and $\lambda_h$ is set to $40$. After 40 epochs, $\lambda_w$ is slowly increased with step size of $0.2$ until it reaches the max value of $2$. 

Both local and world correspondence prediction networks are initially trained using teacher forcing \cite{lamb2016professor}. For local network, a sampling probability is set for using the ground truth or predicted bounding box for feature extraction. Ground truth features are chosen with higher probability in the initial epochs and and slowly decreased until only predicted bounding box is used. Similarly, for world correspondence network the probability is set between ground truth local correspondences and predicted local correspondences, where ground truth local correspondences are chooses initially with higher probability. 

Overall we train the network for 300 epochs; PyTorch is used in constructing and training the network with Adam \cite{kingma2014adam} optimizer and an initial learning rate of 1e-4 with step learning rate decay of step size $20$ and $\gamma=0.9$. The weights of the network are initialized by He initialization \cite{he2015delving}. 

\textbf{Datasets}.
We train our model on the total segmentator dataset \cite{wasserthal2023totalsegmentator} that consists of 1188 CT scans and is publicly available. 
We select seven anatomies: heart ventricle left (HVL), heart ventricle right (HVR), heart atrium left (HAL), heart atrium right (HAR), lung upper lobe left (LULL), lung upper lobe right (LULR) and spleen (S). From the  total segmentator dataset we only select the images and have these anatomies and discard the instances with improper segmentations. 
We divide the dataset into training (624 samples), validation (62 samples) and test (63 samples) sets and use ShapeWorks \cite{cates2017shapeworks} to form the initial PDM with 1024 points from training and validation set for each anatomy individually. 


\textbf{Evaluations}: As we are the first to propose both the delineation of anatomy in image space and simultaneous predictions for multiple anatomies, there are no relevant comparisons. However, we compare our results with {\sc DeepSSM} \cite{bhalodia2018deep} (proposed for getting shape representation of single anatomy directly from already cropped and aligned images), for both local and world correspondences prediction.

{\sc DeepSSM-L} refers to the model trained individually on full images i.e. uncropped and unaligned images to regress local particles. Additionally, for local correspondences, we train UNet as a baseline for each anatomy individually and use the predicted segmentations to reconstruct the mesh and compare the surface-to-surface distance. This measures whether the reconstructions resulting from the shape model of our method provide better shape information as compared to segmentation.

For world correspondences, we train two variations of  {\sc DeepSSM}, {\sc DeepSSM-W}, and {\sc DeepSSM-WF}. {\sc DeepSSM-WF} uses full images (uncropped and unaligned) to individually learn shape representation for each anatomy.  {\sc DeepSSM-WF} is the closest comparison to our proposed methodology because this model is trained on full images without any cropping of the anatomy of interest but still trained on a single anatomy. For {\sc DeepSSM-W}, the images are first cropped and aligned as a pre-processing step, and then {\sc DeepSSM} model is used for training shape representation for each anatomy.

\textbf{Metrics}: For each organ, we evaluate the performance of each model by comparing the predicted particles to their ground truths and computing the root mean squared error (RMSE) averaged over each dimension. We also reconstruct the mesh from the predicted and ground truth particles and evaluate the surface-to-surface distance. 

\subsection{MASSM Results}



\begin{figure}
    \centering
    \includegraphics[width=1.0\linewidth]{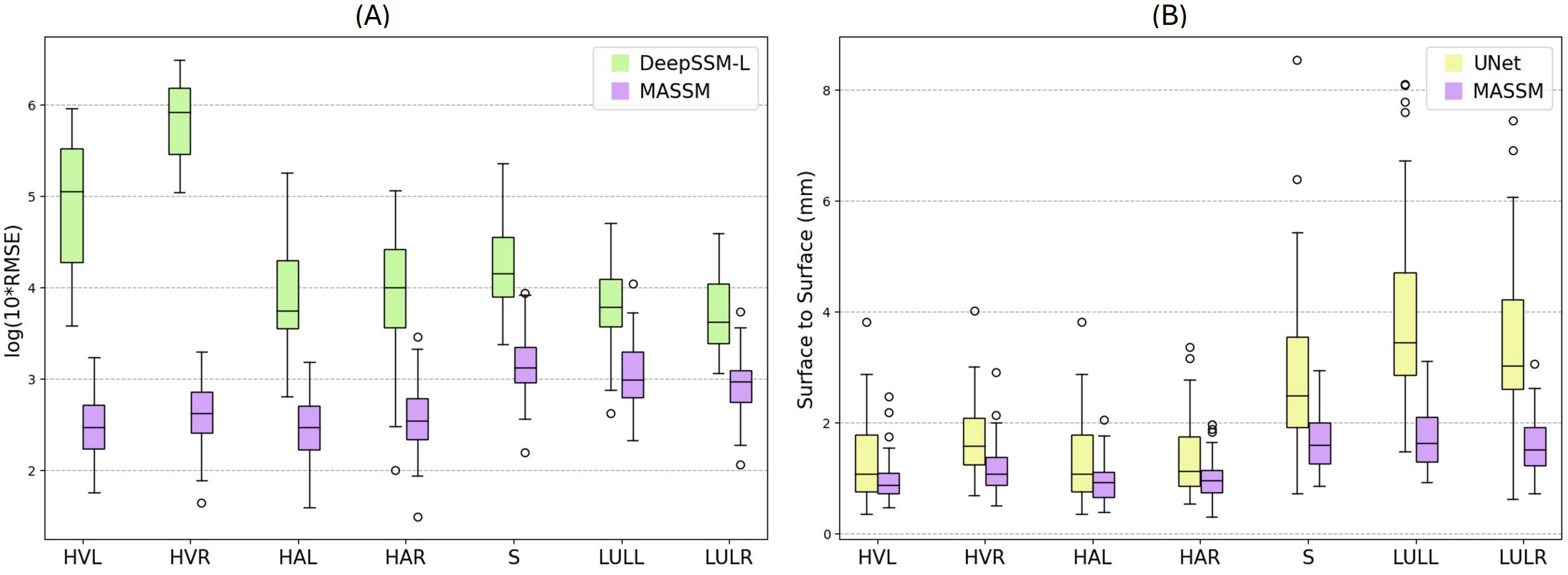}
    \caption{\textbf{Local Correspondence Prediction Results.} (A) RMSE on local particles reported for all seven anatomies under test when compared with {\sc DeepSSM-L}\cite{bhalodia2018deepssm}.(B) Surface-to-surface distance (mm) for local particles reconstructions reported for seven anatomies.}
    \label{localres}
\end{figure}

\textbf{Local Correspondence Prediction Results.} Figure \ref{localres}A shows the box plot illustrating the RMSE and Figure \ref{localres}B shows surface-to-surface distance box plot for the reconstructed mesh on the test data. Our proposed end-to-end framework outperforms {\sc DeepSSM-L} on the RMSE metric by a large margin. Predicting local particles poses a considerable challenge, since it necessitates consideration of rigid transformations. Our method adeptly addresses this challenge through a staged approach. Initially, we pinpoint the center and solve for translation. Furthermore, the incorporation of a template provides a valuable prior over shape. Post alignment, the model efficiently tackles minor variations in a constrained setup, contributing to the observed performance boost. 

From Figures \ref{localres}A and \ref{localres}B, we can observe that MASSM consistently outperforms the UNet segmentations for all anatomies when comparing surface-to-surface mesh distances. Unlike segmentation, which involves pixel-wise classification and often results in noisy labels, SSM encodes a statistical prior that ensures densely packed correspondences resulting in robust surface reconstructions. Consequently, our method demonstrates reduced sensitivity to outliers and provides more accurate and reliable shape information.

\begin{figure}[!htb]
  \centering
  \includegraphics[width=0.9\linewidth]{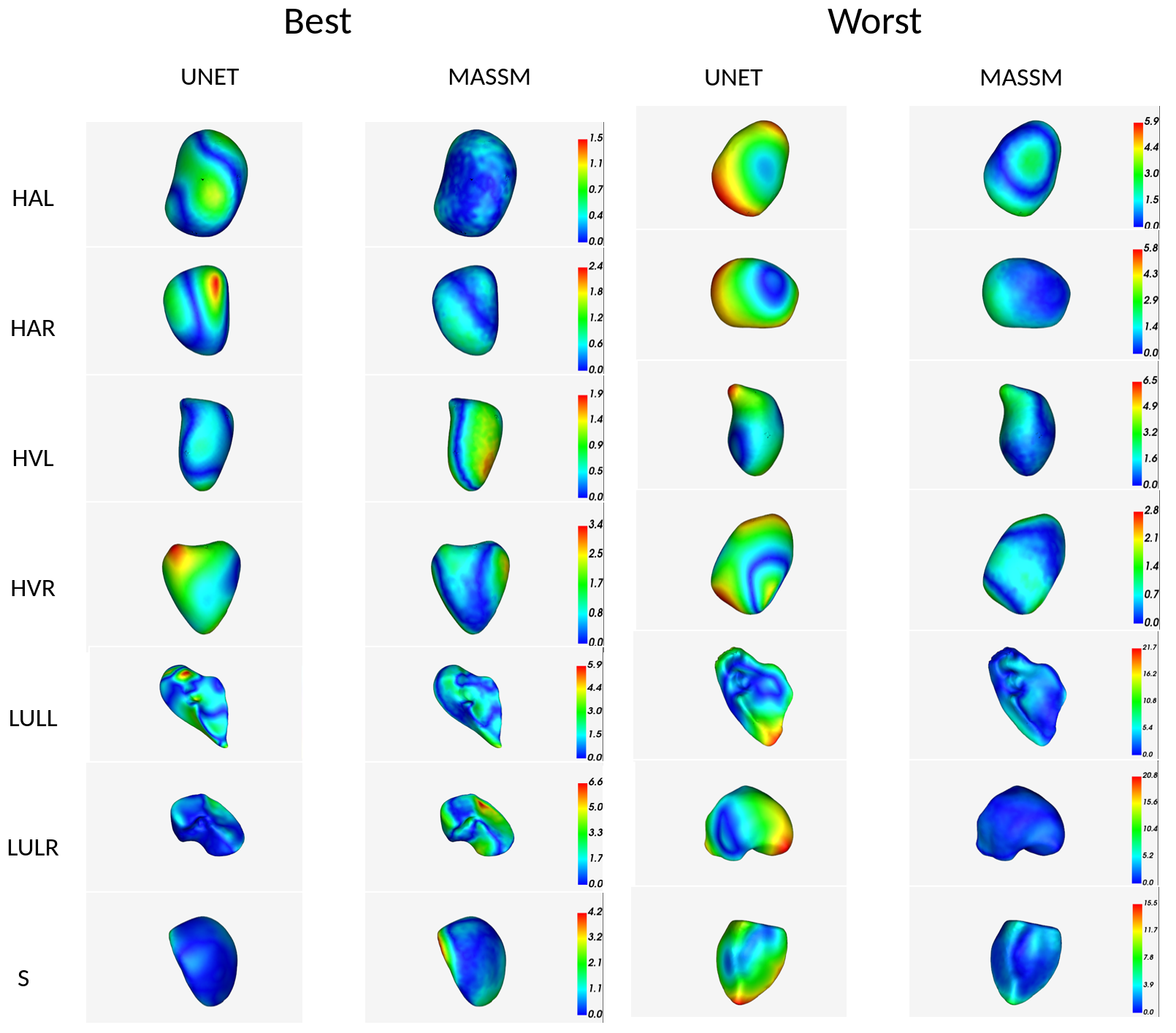}
  \caption{Surface-to-surface distance errors for shape reconstruction of local correspondences, depicted as a heatmap on ground truth reconstructed meshes, showcasing best and worst-case scenarios across 7 anatomies. Heart ventricle left (HVL), heart ventricle right (HVR), heart atrium left (HAL), heart atrium right (HAR), lung upper lobe left (LULL), lung upper lobe right (LULR) and spleen (S).}
  \label{sup_l}
  \vspace{-2em}
\end{figure}

We provide the surface-to-surface distance visualization of reconstructed mesh for best and worst cases for the test set for local correspondences in Figure \ref{sup_l}. Figure \ref{sup_l} reinforces the above observation, as even in the best-case scenario for UNet, outliers are present, whereas MASSM consistently offers more uniform reconstructions. 

\begin{figure}
    \centering
    \includegraphics[width=1.0\linewidth]{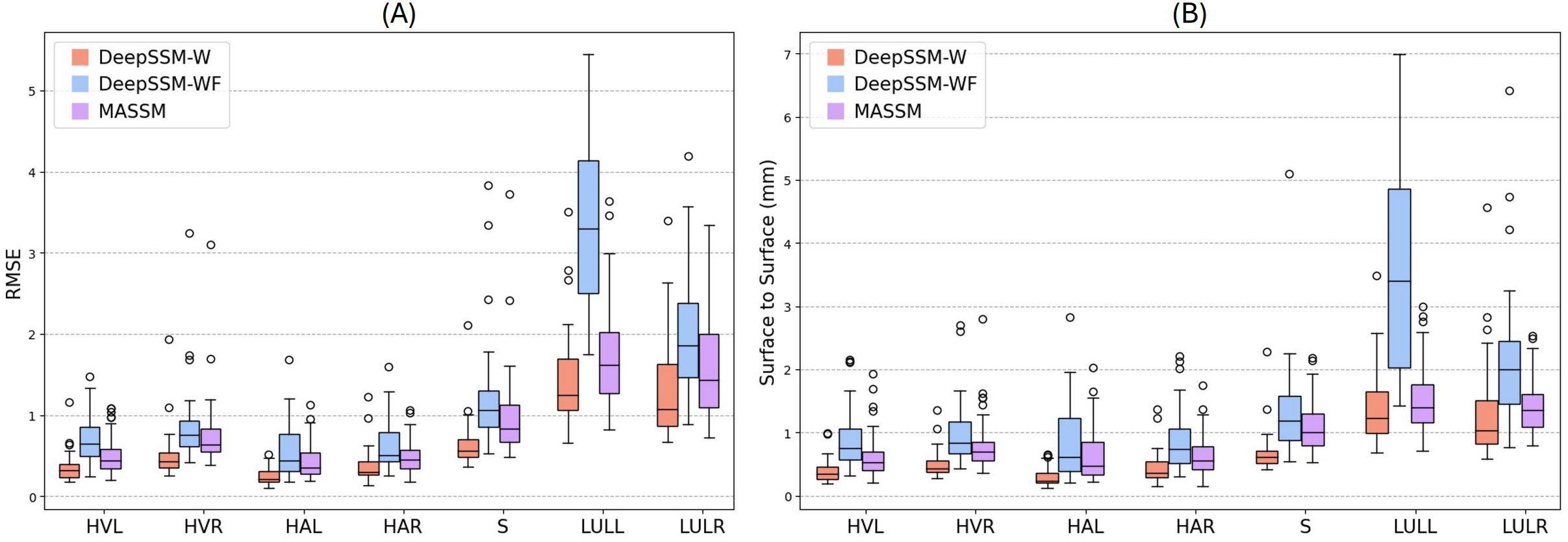}
    \caption{\textbf{World Particle Prediction Results.} (A) RMSE on World Particles reported for seven anatomies. (B) Surface-to-surface distance (mm) for world particles reconstructions reported for seven anatomies. }
    \label{wres}
\end{figure}

\vspace{10pt}
\textbf{World Correspondences Prediction Results.} The box plots for test data representing the RMSE and surface-to-surface distance for reconstructed mesh for world particles are shown in Figure \ref{wres}A and Figure \ref{wres}B respectively. In contrast to {\sc DeepSSM-W}, our method exhibits slightly lower performance, we attribute it to the fact that {\sc DeepSSM-W} was trained on cropped and aligned data specific to each anatomy, and therefore this baseline serves as the upper bound on expected MASSM performance.

\begin{figure}[!htb]
  \centering
  \includegraphics[width=1.0\linewidth]{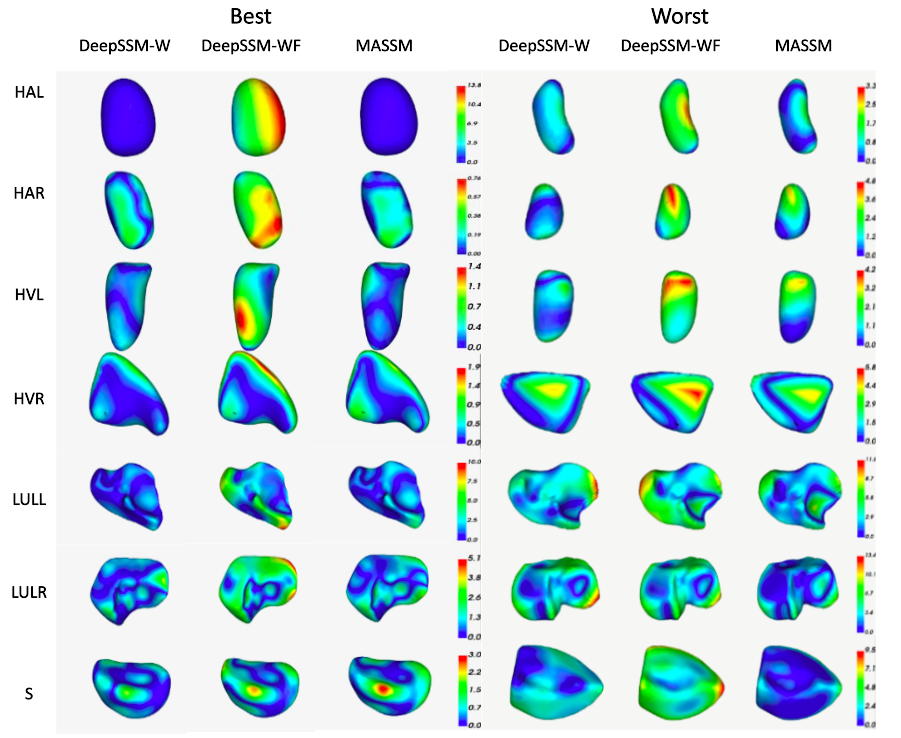}
  \caption{Surface-to-surface distance errors for shape reconstruction of world correspondences, depicted as a heatmap on ground truth reconstructed meshes, showcasing best and worst-case scenarios across 7 anatomies. Heart ventricle left (HVL), heart ventricle right (HVR), heart atrium left (HAL), heart atrium right (HAR), lung upper lobe left (LULL), lung upper lobe right (LULR) and spleen (S).}
  \label{sup_w}
\end{figure}

Our proposed method consistently outperforms {\sc DeepSSM-WF} for all anatomies under test. We attribute our method's higher performance as compared to {\sc DeepSSM-WF} to its reliance on local correspondences instead of image features. Once local particles are identified, the subsequent task involves aligning them to the template for world correspondence. This alignment step is comparatively easier to learn than extracting features from full images that correspond to world shape information. Figure \ref{sup_w} depicts the surface reconstruction heatmap for different anatomies. Notably, {\sc DeepSSM-W} demonstrates the most favorable performance, followed by MASSM and then {\sc DeepSSM-WF}.

\subsection{Ablation}

\begin{figure}[!htb]
  \centering
  \includegraphics[width=1.0\linewidth]{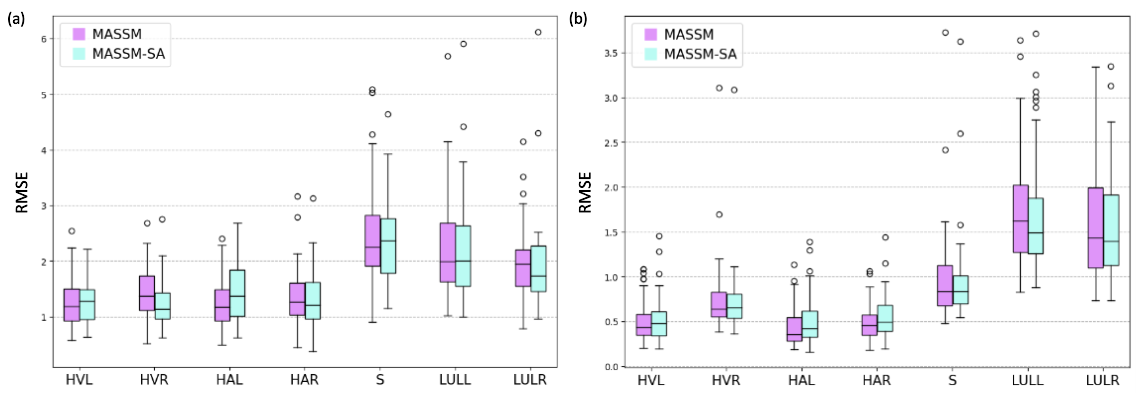}
  \caption{RMSE for both (a) Local particles, and (b) World particles, when comparing to multi vs single anatomy architecture.}%
  \label{a1}
\end{figure}

To further investigate the impact of our multi-anatomy shape representation, we conducted an ablation study by training our network MASSM independently for each anatomy. 
This analysis was conducted to gain deeper insights into whether the improved performance of MASSM is attributable to its multi-anatomy shape representation prediction or the multi-stage approach employed for detecting local particles. Results are shown in Figure \ref{a1}. MASSM-SA denotes MASSM trained for each anatomy individually, we observe that the RMSE values for both local and world correspondences are nearly identical. The results suggest that the improved performance of MASSM may be attributed more to the multi-stage approach rather than the prediction of multi-anatomy shape representation. The diversity of shapes (heart, lung, spleen) in the dataset likely diminished the impact of the multi-anatomy shape representation. 
In future work, we will apply MASSM to more uniform shapes, such as various vertebrae of the human spine, where the underlying structures are relatively similar for different vertebrae.

While predicting multi-anatomy shape representation does not enhance performance, it significantly reduces training time. Notably, training MASSM for 300 epochs takes 2 days, whereas training MASSM-SA individually takes 1 day per anatomy, totaling 7 days. Therefore, although MASSM shows no advantage in performance, its efficiency in handling multiple anatomies via a single model is markedly superior. 

\vspace{-1em}

%% file: conclusion.tex
\section{Conclusion}
MASSM is a end-to-end deep learning framework designed to extract shape representations for multiple anatomies within an image simultaneouly. MASSM is the first model that predicts both local and world correspondences for multiple anatomies. This approach for deep learning-based shape modeling concurrently identifies the anatomy of interest while predicting its corresponding population and image level statistical shape representations. 
Additionally, our method eliminates the need for the manual preprocessing of input images required by other shape modeling techniques \cite{bhalodia2018deep,bhalodia2018deepssm,adams2020uncertain,adams2022images}. In comparison to the baseline DeepSSM models, our proposed model exhibits qualitative and quantitative performance that is greater than the baselines for world correspondences. Additionally, it demonstrates superior results for local correspondences, even when comparing surface-to-surface distances with UNet. 
Our findings underscore the significance of local particles in providing superior shape information compared to segmentation, making them indispensable for medical imaging tasks. Additionally, our findings emphasize the training efficiency of the multi-anatomy architecture over single anatomy, providing comparable performance but with significantly lower training time. In the future, we aim to enhance scalability of \name by increasing the number of anatomies modeled and reducing training time. We aim to further explore efficient deep learning architectures to reduce the computational load for end-user applications.
\vspace{-1em}
\section*{Acknowledgements}
\vspace{-1em}
The National Institutes of Health supported this work under grant numbers NIBIB-U24EB029011 and NIAMS-R01AR076120. The content is solely the authors' responsibility and does not necessarily represent the official views of the National Institutes of Health.